\documentclass[review,11pt]{elsarticle}
\usepackage{lineno}
\usepackage[colorlinks,bookmarksopen,bookmarksnumbered,citecolor=blue,linkcolor=magenta]{hyperref}
\usepackage{gensymb}
\usepackage{xspace}
\usepackage{amsmath,amssymb}
\usepackage{algorithm2e}
\usepackage{verbatim}
\newcommand{\penal}{Pen}
\usepackage{url}

\newcommand{\postVRP}{PostVRP\xspace}
\newcommand{\depo}{\pi}
\newcommand{\nbb}{\mathbb{N}}
\newcommand{\sol}{\mathcal{S}}
\newcommand{\rota}{\mathcal{R}}
\newcommand{\particao}[1]{Partition(#1)}
\newcommand{\compri}[1]{W(#1)}
\newcommand{\rotamax}{\rota_{max}}
\newcommand{\R}{\mathbb{R}}
\newcommand{\Pol}{\mathcal{P}}
\newcommand{\artur}{Artur Nogueira\xspace}
\newcommand{\largura}{wth}
\newcommand{\dens}{D}
\newcommand{\peso}{Prob}
\newcommand{\location}{loc}
\newcommand{\cro}{cross}
\newcommand{\mpath}{minpath}
\newcommand{\atra}{R\xspace}
\newcommand{\atrana}{central\xspace}
\newcommand{\atranb}{peripher\xspace}
\newcommand{\atranc}{distant\xspace}
\newcommand{\atrand}{isolated\xspace}
\newcommand{\atrb}{T}
\newcommand{\atrbna}{avenue}
\newcommand{\atrbnb}{street}
\newcommand{\atrbnc}{way}
\newcommand{\atrbnd}{highway}
\newcommand{\atrc}{Z}
\newcommand{\atrcna}{commercial}
\newcommand{\atrcnb}{mixed}
\newcommand{\atrcnc}{residential}
\newcommand{\numeroRuas}{422\xspace}
\newcommand{\ruas}{Streets}
\usepackage{caption}

\newtheorem{definicao}{Definition}

\modulolinenumbers[5]

\journal{~}

\biboptions{authoryear}

\begin{document}
\captionsetup[figure]{labelfont={bf},name={Fig.},labelsep=period}
\captionsetup[table]{labelfont={bf},name={Table},labelsep=period}
\begin{frontmatter}

\title{Multi-Objective Vehicle Routing Problem Applied to Large Scale Post Office Deliveries
}

%
%
%



\author[mymainaddress]{Luis A. A. Meira\corref{cor1}}
\ead[url]{http://www.ft.unicamp.br/~meira}
\ead{meira@ft.unicamp.br}

\cortext[cor1]{Corresponding author}

\author[mymainaddress]{Paulo S. Martins}
\ead{paulo@ft.unicamp.br }

\author[mymainaddress]{Mauro Menzori}
\ead{mauro@ft.unicamp.br}

\author[mymainaddress]{Guilherme A. Zeni}
\ead{g146284@dac.unicamp.br}

\address[mymainaddress]{School of Technology, University of Campinas,\\ Paschoal Marmo, 1888, Limeira, SP, Brazil}

\begin{abstract}
This template helps you to create a properly formatted \LaTeX\ manuscript.
\end{abstract}

\begin{keyword}
\texttt{elsarticle.cls}\sep \LaTeX\sep Elsevier \sep template
\MSC[2010] 00-01\sep 99-00
\end{keyword}

\begin{abstract}
The number of optimization techniques in the combinatorial domain is large and diversified.
 Nevertheless, real-world based benchmarks for testing algorithms are few.
 This work creates an extensible real-world mail delivery benchmark
to the Vehicle Routing Problem (VRP) in a planar graph embedded in the 2D Euclidean space. 
Such problem is {multi-objective} on a roadmap with up to 25 vehicles and {30,000 deliveries per day}.
%
Each instance models one generic day of mail delivery, 
allowing both comparison and validation of optimization algorithms for routing problems. 
The benchmark may be extended to model other scenarios.
\end{abstract}

\begin{keyword}
Routing \sep
Large scale optimization \sep
Multi-objective optimization \sep
Logistics \sep
OR in service industries. 
\end{keyword}

\end{frontmatter}


\section{Introduction}
 
Benchmarks are found in various fields of science, such as geology~\citep{correia2015unisim}, economics~\citep{jorion1997value}, and climatology~\citep{tol2002estimates}, among other areas. 
They play {a central role} in computer science, e.g., in image processing~\citep{du1990texture,krizhevsky2012imagenet,huang2007labeled}, hardware performance~\citep{che2009rodinia}, and optimization~\citep{reinelt1991tsplib,kolisch1997psplib,burkard1997qaplib}.

Within the context of optimization,~\cite{johnson2002theoretician} divided algorithm analysis in three approaches: the worst-case, the average-case, and the experimental analysis. Regarding experimental papers, he identifies four cases: \textit{(i)} solving a real problem;
 \textit{(ii)} providing evidence that one algorithm is superior to others; \textit{(iii)} better understanding a problem; and \textit{(iv)} studying the average case.
He proposes the use of well-established benchmarks to provide evidence of the superiority of an algorithm (item \textit{ii}). Such papers are called \textit{horse race papers}.

Johnson highlights that reproducibility and comparability are essential aspects of any experimental paper. 
The author mentions the difficulty in justifying experiments on problems with no direct application. Such problems have no real instances and the researcher is forced to generate the data in a \textit{vacuum}.

Our work deals with a variant of the Vehicle Routing Problem (VRP) based on a real mail delivery case in the city of \artur. 
The post office receives thousands of mail items to be delivered on a daily basis.
 Such mail is distributed to a set of 15-25 mail carriers for on-foot delivery. Each mail carrier is modeled as a vehicle and each delivery point is a customer. 
 This variant is named here as Post Office Deliveries VRP (\postVRP).
 
Domain experts indicate that the \postVRP has three main objective functions to be minimized (while maintaining the feasibility of the solutions):
\textit{(i)}~ route length; 
\textit{(ii)} unfairness, measured as the workload (i.e. route length) variance among the mail carriers; 
and \textit{(iii)} number of mail carriers.

The \postVRP considers uncapacitated vehicles and constrained route length. Each mail carrier is allowed to carry a maximum load from 8 to 10 kg. A support truck restocks the mail carriers turning their capacities unlimited. Each mail carrier must follow a 6-to-8-hour working day, which implies a maximum capacity for the route length.

A limited route length is a constraint that models several real-world cases: A helicopter has a route length limited by the capacity of its fuel tank. Workers, in general, have a time window to operate the vehicle, which likewise limits the length of the route.



\textbf{ Contribution:} 
This work presents a case study modeled in a Brazilian city located at $22\degree 34' 22" S 47\degree 10' 22" W$. 
The proposed benchmark contains up to 30,000 customers.
We make available the benchmark tool so that it is possible to create new arbitrarily large instances. 
The methodology can be applied to other cities as well as to other VRP variants. 



The remainder of this paper is organized as follows: the background and review of relevant work is provided in Section~\ref{sec:prework}; in Section~\ref{sec:not} we introduce the notation and definitions; Section~\ref{sec:tool} presents the model; and Section~\ref{sec:bench} addresses a real-world PostVRP benchmark case. Finally, we summarize and conclude in Section~\ref{sec:conc}.


\section{Literature review} \label{sec:prework}

 One of the first references to the VRP dates back to 1959~\citep{dantzig1959truck}, under the name {Truck Dispatching Problem}, a generalization of the Traveler Salesman Problem {(TSP)}.
The term VRP was first seen in the paper \cite{christofides1976vehicle}. Christophides defines VRP as a generic name, given to a class of problems that involves the visit of ``customers'' using vehicles.
 
Real world aspects may impose variants to the problem.
For example, the {Capacitated-VRP (CVRP)}
considers a limit to the vehicle capacity~\citep{fukasawa2006robust}, the {VRP with Time Windows (VRPTW)}
accounts for the delivery time windows~\citep{kallehauge2005vehicle}, and the {Multi-Depot VRP (MDVRP)} extends the number of depots~\citep{renaud1996tabu}.
 Other variants may be easily found in the literature.
 
\cite{reinelt1991tsplib} created a benchmark for the TSP known as TSPLib. In his work, he consolidated non-solved instances from 20 distinct papers. His repository, named TSPLIB95~\citep{reinelt1995tsplib95}, has instances of both the symmetric and the asymmetric Traveling Salesman Problem (TSP/aTSP) as well as three related problems: \textit{(i)} CVRP; \textit{(ii)} Sequential Ordered Problem (SOP); and~\textit{(iii)} Hamiltonian Cycle Problem (HCP).

The number of instances is 113, 19, 16, 41, 9 for TSP, aTSP, CVRP, SOP, and HCP, respectively. The number of vertices varies from 14 to 85,900 for the TSP, 17 to 443 for the aTSP, 7 to 262 for the CVRP, 7 to 378 for the SOP, and from 1,000 to 5,000 for the HCP.

The optimum of all TSPLib instances was finally achieved in 2007,
after sixteen years of notable progress in algorithm development.
The optimum of the d15112 instance was found in 2001 \citep{applegate2011travelingP506}. This instance contains 15,112 German cities and it required 22.6 years of processing split across 110 500 MHz processors~\citep{waterloosite}. 
The instance pla33810 was solved in March 2004~\citep{applegate2011travelingP506}. The pla33810 instance represents a printed circuit board with 33,810 nodes and it was solved in 15.7 years of processing~\citep{espinoza2006linear}. 
The last instance of the TSPLib, called pla85900, was solved in 2006~\citep{applegate2011travelingP506}. This instance contains 85,900 nodes representing a VLSI application. 

\cite{solomon1987algorithms} created a benchmark for the VRPTW in 1987. It is composed of 56 instances partitioned in six sets. 
The number of customers is 100 in all instances. The vehicle has a fixed capacity and the customers have an integer demand. 
The number of vehicles is not fixed: it derives from the fact that capacity is limited. Under this viewpoint, this can be considered a multi-objective problem. It aims to minimize the route and the number of vehicles. 

\begin{figure}[ht]\centerline{
\includegraphics[width=.5\textwidth]{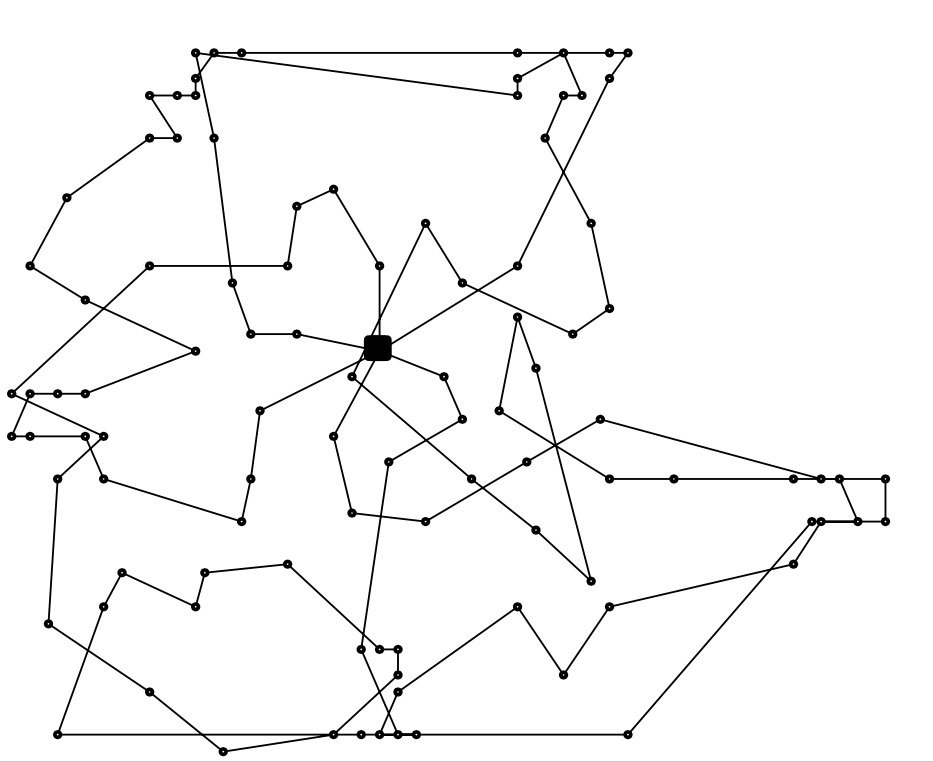}
}
 \caption{The Solomon's RC207 instance composed of 100 customers and one depot with a solution containing three routes~\citep{bent2004two}. The picture does not show the capacity and the time-window constraints.}
 \label{figSolomonInst}
 \end{figure}

The first optimum solution was published by~\cite{kohl19992}. 
\cite{chabrier2006vehicle} solved 17 open instances in the benchmark.
 \cite{amini2010pso} obtained solutions very close to the optimum, considering only the first 25 customers. In July 2015, 28 years after having launched the benchmark, \cite{jawarneh2015sequential} published 
a Bee Colony Optimization metaheuristic. 
Such algorithm reached 11 new best results in Solomon's VRPTW instances.
 It is surprising that such small instances present a quite complex internal structure to be optimized.
Fig.~\ref{figSolomonInst} shows a Solomon's instance composed of 100 customers and a given solution considering three vehicles.
 
Regardless of their complexity, the TSPLib and the Solomon benchmarks have a number of customers between 100 and 262 for the VRP, which is currently a small value.
\cite{gehring1999parallel} extended the Solomon’s instances, 
thus creating a benchmark for the VRPTW with the number of customers varying from 100 to 1,000.


For the CVRP, ABEFMP is a largely used set of instances, in which \cite{augerat1995computational} proposed the A, B, P classes in 1995, and~\citep{christofides1969algorithm,fisher1994optimal,christofides1979vehicle} proposed the E, F, M classes in 1969, 1994 and 1979, respectively. In their benchmark, the number of customers varies from 13 to 200, and the number of vehicles varies from 2 to 17.

\cite{fukasawa2006robust} and \cite{contardo2014new}, among others, obtained the optimum in different ABEFMP instances. \cite{pecin2014improved} found the optimum solution for the last unsolved instance, named \textit{M-n151-k12}, 35 years after its presentation by \cite{christofides1979vehicle}. Despite that, most of those instances are very simple to solve nowadays.

\cite{golden1998impact} proposed new instances for the CVRP. It is a set of 20 instances, with the number of customers varying from 240 to 483. Such benchmark remains entertaining, because most of its instances still do not have an optimum established~\citep{CVRPLIB}.
\cite{li2005very} created a set of instances with the number of customers between 560 and 1200. 
Currently, no optimum has been found for any of the instances~\citep{CVRPLIB}.

\cite{CVRPLIB} created the CVRPLib where they consolidated the CVRP instances of~\citep{augerat1995computational,christofides1969algorithm,christofides1979vehicle,fisher1994optimal,golden1998impact,li2005very}. 
In addition, \cite{uchoa2017} generated new instances with the number of customers between 100 and 1,000. Their work indicates the lack of well-established challenging benchmarks for the VRP.

Uchoa \emph{et al.} also highlight the fact that benchmarks are artificially created. Solomon and Uchoa \emph{et al.} generated their own instances using random points. In the ABEFMP benchmark, some random instances are generated and other instances represent real problems. However, in all the instances the customers are points in the Euclidean space. The instances~\cite{golden1998impact} and~\cite{li2005very} are artificial as well.



\section{Notation and definition} \label{sec:not}

Consider a set of elements $S$
 where 
a $depot$ is a special element $\depo\in S$. This work does not address multiple-depot variants to the VRP.
The set of customers is defined by $C=S\setminus\{\depo\}$ and the number of customers is denoted by $n$, where $C=\{c_1,\ldots,c_n\}$.
The number of vehicles in the fleet is represented by $k\in \nbb$. 
The value $k$ is traditionally considered a constant, but it is possible to define variants to VRP where $k$ is variable.
Let $w:S\times S\rightarrow \mathbb{N}$ be the cost between any two elements in $S$. 
 Let
$\sol(C,k) = (c_1,\ldots,c_n,\pi,\ldots,\pi).$
This sequence is created as follows: \textit{(i)} all elements in $C$ are inserted
in $\sol$; \textit{(ii)} the depot vertex is inserted $k-1$ times.

\begin{figure}[htb]
\begin{center}
\includegraphics[width=6cm]{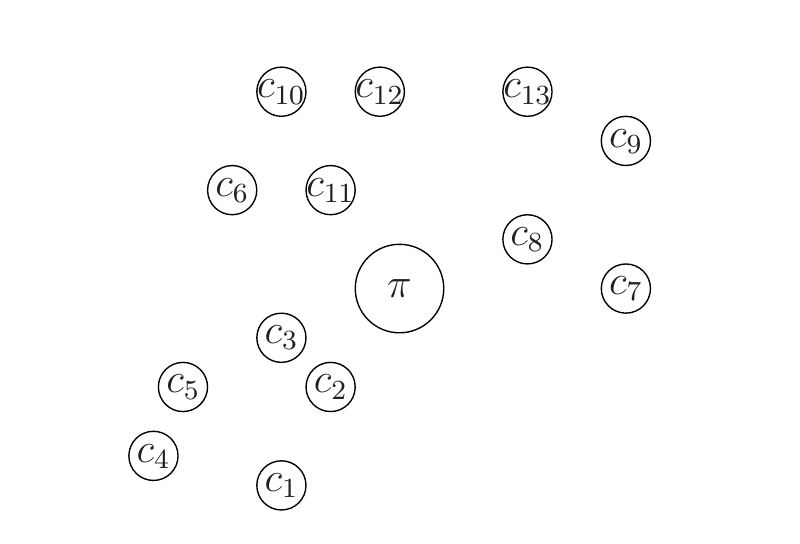}
\includegraphics[width=6cm]{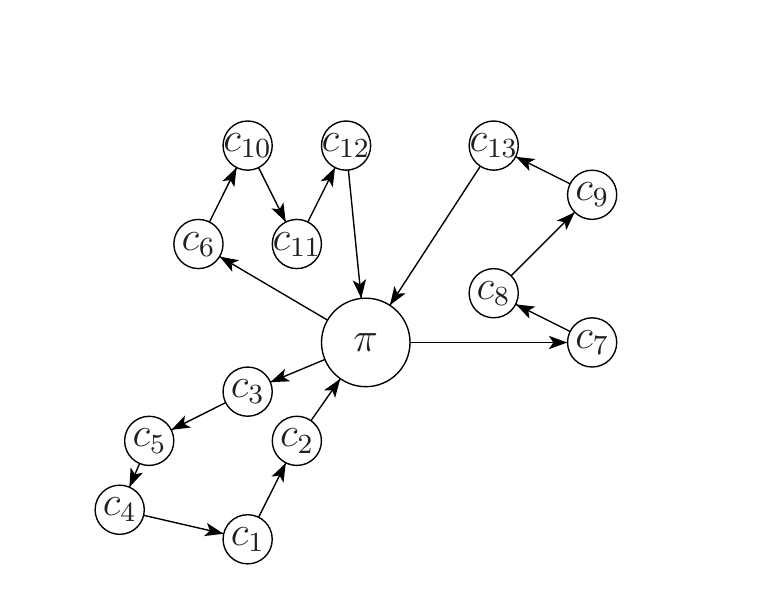}
\end{center}
\caption{Sample VRP vertices (left) and solution $\sol'=(c_3,c_5,c_4,c_1,c_2,\mathbf{\pi},c_6,c_{10},c_{11},c_{12},$ $\mathbf{\pi},c_7,c_8,c_9,c_{13})$ 
(right).
}
\label{fig:instance}
\end{figure}

Each permutation of $\sol(C,k)$ represents a solution to the VRP. 
For example, consider the graph 
and the vertices described in Fig.~\ref{fig:instance}, and suppose that the number of vehicles is three (i.e. $k=3$).
The $\sol(C,3)$ sequence is given by $\sol(C,3)=(c_1,\ldots,c_{13},\pi,\pi).$
For example, the permutation
$\sol'=(c_3,c_5,c_4,c_1,c_2,\pi,c_6,c_{10},c_{11},c_{12},\pi,c_7,c_8,c_9,c_{13})$
is the solution described in Fig.~\ref{fig:instance}.

All routes begin and end at the depot.
The $\sol'$ solution represents a partition of the clients in three routes: $\rota_1=(c_3,c_5,c_4,c_1,c_2)$, 
$\rota_2=(c_6,c_{10},c_{11},c_{12})$ and $\rota_3=(c_7,c_8,c_9,c_{13})$.
The vertex $\depo$ is used to create a partition of the sequence in $k'\leq k$ routes.
 Let $\displaystyle\particao{\sol}=\left(\rota_1,\ldots,\rota_{k'}\right)$.
 By definition, empty routes are not part of $\particao{\sol}$. Thus, $\particao{1,2,\depo,\depo,3,4}$ is $\{(1,2),(3,4)\}$ and not $\{(1,2),(),(3,4)\}$, 
 i.e. $k'\leq k$.

The length of a route $\rota=(r_1,\ldots,r_m)$ is given by:
$$\displaystyle\compri{\rota}=w(\depo,r_1)+w(r_m,\depo)+\sum_{i=1}^{m-1}w(r_i,r_{i+1}).$$
The length of a solution $\sol=(s_1,\ldots,s_m)$ is calculated as: 
$$\displaystyle\compri{\sol}=w(\depo,s_1)+w(s_m,\depo)+\sum_{i=1}^{m-1}w(s_i,s_{i+1}).$$

The sequence $\sol$ contains edges between deliveries and the depot, other than the first and last edge that need to be included in $\compri{\sol}$.
The number of vehicles used in a given solution is equal to the number of non-empty routes $|\particao{\sol}|.$
If the number of vehicles is $k$ and non-empty routes are allowed, we have the constraint $|\particao{\sol}|=k$.
If the number of vehicles are at most $k$, or if empty routes are allowed, we have $|\particao{\sol}|\leq k$.
If the number of vehicles is not a part of the input, the domain may be given by the permutation of the 
$\sol(C,k)$ sequence. In this case, the number of vehicles is defined during the optimization step.

Given a feasible solution, it is necessary to calculate its costs.
The most common objective function to be minimized is the length of the
solution: $$\displaystyle f_1(\sol)=\compri{\sol}.$$
Another objective function consists in finding a feasible solution that minimizes
 the number of vehicles: $$\displaystyle f_2(\sol)=|\particao{\sol}|.$$ Suppose there are 25 available 
 mail carriers and a feasible solution with 21 routes. In this case, the post office may allocate the four available mail carriers
 to other internal tasks.

Finally, it is required that the solution meet the fairness criteria, i.e.
routes should be assigned in a way that balances out the workload (route length) 
among the mail carriers.
The way we modeled fairness was through minimizing the variance of the route lengths:
$$\displaystyle f_3(\sol)=\sqrt{\frac{\displaystyle\sum_{\rota\in\particao{\sol}}\left(\compri{\rota}-\overline{\compri{\rota}}\right)^2}{|\particao{\sol}|-1}}.$$

VRP is a set of problems that consists of visiting customers using vehicles.
Each variant has additional feasibility constraints,
such as not allowing empty routes ($|\particao{\sol}|=k$).
The \postVRP assumes that the length of the route is limited.
Let $\rotamax$ be the maximum allowed route length, i.e.
$W(\rota)\leq \rotamax.$

\begin{definicao}[PostVRP] Given a set of elements $S$, a weight function $w:S\times S\rightarrow \mathbb{N}$, a constant $k\in\mathbb{N}$ representing the maximum number of vehicles, a special vertex
$\pi\in S$ and a maximum route length $\rotamax\in \mathbb{N}$. 
Let $C\gets S\setminus\{\pi\}$. Consider the sequence $\sol(C,k)$, and let $Pe$ be the set of all feasible permutations of $\sol(C,k)$ with
respect to $\rotamax$. The $\postVRP$ problem consists of minimizing 
$\displaystyle\left(f_1(\sol),f_2(\sol),f_3(\sol)\right)$ subject to $\sol\in Pe$.
\end{definicao}



\section{Model description} \label{sec:tool} 

This section describes the \postVRP model.
We start the process of creating the benchmark by mapping each street onto a street map graph.
Each street $St$ is modeled as a polygonal chain, which is defined as a set of planar coordinates. For example,
suppose a University St. modeled as a polygonal chain $P=(c_1,\ldots,c_{n'})$,
where $c\in\R^2$ for all $c\in P$. 
The complete map graph has a set $\Pol=(P_1,\ldots,P_{n''})$ of polygonal chains, one for each street.

We create a graph $G(V,E)$ based on $\Pol$.
Each vertex $v\in V$ is associated with a Cartesian coordinate $(x_v,y_v)\in \mathbb{R}^2$
 and each edge $e=(u,v)$ is a straight line segment between $u$ and $v$.
 The edge weight is $w'(e)=\sqrt{(x_u-x_v)^2+(y_u-y_v)^2}$.
The vertices associated with corners are automatically built by a line segment intersection algorithm.

\begin{figure}[ht!]
	\includegraphics[width=0.45\textwidth]{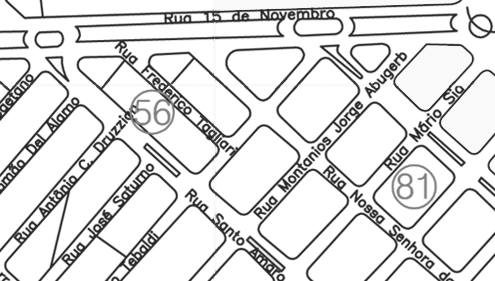} ~~
	\includegraphics[width=0.45\textwidth]{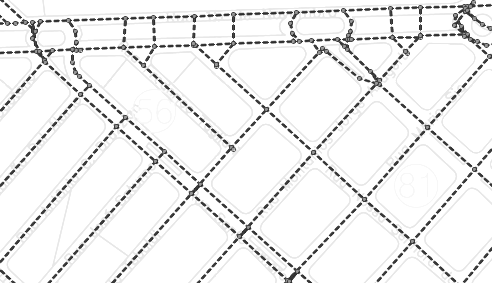}
 	\caption{
	A section of the city map (left). Edges and vertices created over the city map (right).}
	\label{mapa}
\end{figure}

 Fig.~\ref{mapa} (right) displays simple streets modeled as one polygonal chain and streets with
islands, which are modeled by two parallel polygonal chains.
 Additional segments are added to allow shortcuts in footpaths.

Given an edge $e$, its street is denoted by $St(e)$.
Each street $St$ has an arbitrarily defined width $\largura(St)\in \mathbb{R}^+$
 that represents the cost of crossing the street. 
{The value $\largura(St)$ can be set to zero, thus
 resulting in no cost to cross the street.}

A non-normalized probability density $\dens(St)$ is assigned to each street. 
 The probability of a street receiving a delivery workload per unit length is directly proportional to the density value
 $\dens$. 
{ Such value is used to create a central street with a large workload compared to 
a distant one.}
The probabilities are outlined in the next subsection.

\subsection{Generating delivery points}

Consider a street map graph $G(V,E)$. The probability of one delivery being assigned to an edge $e$, denoted by
$\peso(e)$, is:
$$ \peso(e) = \frac{D(St(e))w'(e)}{T} \mbox{\rm,~where~} T=\sum_{\forall e'} D(St(e'))w'(e').$$ 
$\peso(e)$ is directly proportional to the edge length $w'(e)$ and to the probability density $\dens(St(e))$,
and it must be normalized to obtain $\sum_{e\in E}\peso(e)=1$.

The location of a given delivery $d$, denoted by $\location(d)$, is composed of three attributes: 
an edge $(u,v)$, a value $\alpha \in[0,1]$ and a label $street\_side=\{\oplus,\ominus\}$. 
 The delivery is positioned at the affine combination 
of $u$ and $v$ in respect to~$\alpha$, i.e. $(x_u,y_u)(\alpha)+(1-\alpha)(x_v,y_v)$.
The street of a delivery $d=(e,\alpha,s)$, denoted by $St(d)$, is the street of the edge $St(e)$.

An integer $n$ represents the number of deliveries,
and an artificial delivery $d_\pi$ is created 
for the depot. The value of $\alpha$ is randomly generated within the interval [0, 1]. The street
side label is an equiprobable random choice in the set $\{ \oplus, \ominus\}$.
Algorithm~\ref{alg:calckp} is then used to create the delivery set.
Given that the number of deliveries is a part of the input,
it is possible to set arbitrarily large instances.

\begin{algorithm}[ht!]
\linespread{1}
\scriptsize
\LinesNumbered
\SetAlgoLined
\KwIn{An integer $n$ and a set of edges $E$ with probabilities $\peso(e)$, $\forall e\in E$}
\KwOut{A set of deliveries $Del$.}
$Del\gets \emptyset$\\
Partition all edge probabilities in the interval $[0,1]$\\
\For{i=1 to n}{
Select a random value $r\in[0,1]$\\
\If{\mbox{\rm $r$ is in the interval associated with $\peso(e)$}}
{
Select a random value $\alpha\in[0,1]$\\
Select a random street side value $s \in\{ \oplus, \ominus\}$\\
$Del \gets Del \cup \{(e,\alpha,s)\}$
}
}
{ 
 \Return{$Del$}
}
\caption{Algorithm to create deliveries.} 
\label{alg:calckp}
\end{algorithm}


\subsection{Defining the weight between a pair of deliveries} 

 The street map graph $G(V,E)$ is used to compute the weight between deliveries.
Given two deliveries $d_a$ and $d_b$, the cost to cross the street is defined as:
$$\cro(d_a,d_b)=\left \{\begin{tabular}{lp{7cm}} $\largura(St(d_a)),$ & if $(d_a,d_b)$ $side$ labels are $\left\{( \oplus, \ominus),(\ominus, \oplus)\right\}$ and $St(d_a)=St(d_b)$, \\ 0, & otherwise. \end{tabular}\right .
$$

{A constant $\beta\in \R^+$, that represents an additional fixed cost per delivery, must be defined.}
The weight between two deliveries $d_a=(e_a,\alpha_a,s_a)$ and $d_b=(e_b,\alpha_b,s_b)$ is given by $w(d_a,d_b)$.
If $e_a=e_b$ then: $$w(d_a,d_b)=|\alpha_a-\alpha_b|w'(e_a)+\cro(d_a,d_b)+\beta.$$

\begin{figure}[htb]
\centerline{\includegraphics[width=7cm]{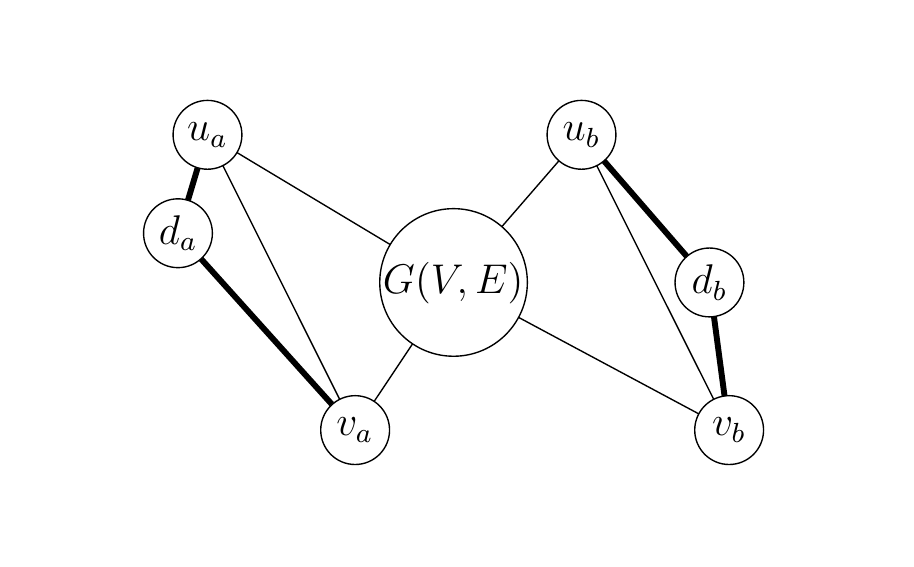}}
\vspace{-1cm}
\caption{The cost between two deliveries $d_a$ and $d_b$ located at distinct edges $e_a$ and $e_b$, respectively.}
\label{fig:w}
\end{figure}

Let $G(V,E)$ be the original street map, $e_a=\{u_a,v_a\}$, $e_b=\{u_b,v_b\}$, and let $G^*(V^*,E^*)$ be defined as:
$V^*=V\cup\{d_a,d_b\}$, $E^*=E\cup\{(u_a,d_a),(v_a,d_a),(u_b,d_b),(v_b,d_b)\}$  (Fig.~\ref{fig:w}),
thus:
$$
w(d_a,d_b)= \mpath(d_a,d_b,G^*)+\cro(d_a,d_b)+\beta.
$$
 The instance is composed of a matrix $w_{n\times n}$, an integer $k$, and an integer $\rotamax$. The first delivery represents the depot.

\subsection{The benchmark tool}

This subsection describes the tool that creates the benchmark. 
It has three configuration files, \emph{background.png},
\emph{model.txt} and \emph{instances.txt}.
The background file contains an image used to improve visualization 
and its resolution is used as the base for the model. 
The model file must contain the following information:
\begin{itemize}
\item Depot location: the coordinate position reference to the depot;
\item Additional cost per delivery: cost to hand out the delivery;
\item Decimal precision: number of digits after the fractional part;
\item Pixel value: value used to convert a pixel into other units;
\item Attributes: attributes used to compute the street probability density and the cost to cross the street;
\item Roadmap: the description of streets including the polygonal chain.
\end{itemize}

 Consider a white background with $500\times500$ pixels and the model described in Fig.~\ref{table:mod}.
The tool will process the model file and create a roadmap (Fig.~\ref{table:map}). The depot is positioned at the closest edge. The probability density $D$ of the 4th Av is 0.4 because
it is a {\scriptsize $[AVE,PERIPHERAL,RADIOACTIVE]$} with values $\{20,0.2,0.1\}$.

\begin{figure}[htb]
\linespread{1}
\begin{tabular}{p{4cm}p{4cm}}
\scriptsize
\verbatiminput{modelA.txt} &
\scriptsize
\verbatiminput{modelB.txt} 
\end{tabular}
\vspace{-.5cm}
\caption{Example of a model file.}
\label{table:mod}
\end{figure}

\begin{figure}[htb]
\begin{center}
\begin{tabular}{p{5.5cm}p{5cm}}
\includegraphics[width=5cm]{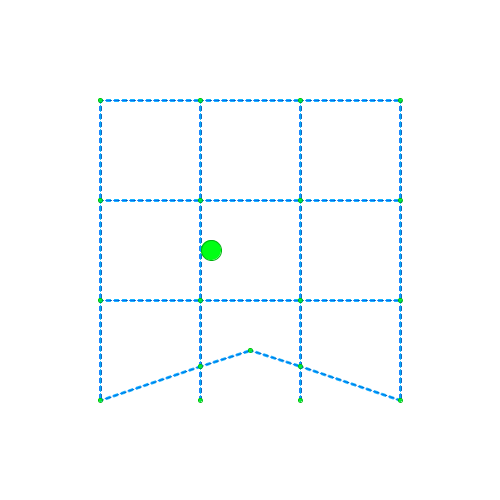}
& 
\scriptsize
\vspace{-4cm}
\begin{tabular}{ccc}
\hline
 St & \dens(St) & \largura(St) pixel\\\hline
1th STR &10&20\\
2th STR &2&20\\
3th STR &1&20\\
4th STR &0.2&20\\
1th Av &20&10\\
2th Av &4&10\\
3th Av &2&10\\
4th Av &0.4&10\\\hline
\end{tabular}
\end{tabular}
\end{center}
\vspace{-.5cm}
\caption{Map based on Fig.~\ref{table:mod} model. }
\label{table:map}
\end{figure}

The last file is named \emph{instance.txt} (Table~\ref{instance:table}). Each line corresponds to an instance in the benchmark. It must contain the instance ID, the directory and subdirectory, the maximum number of vehicles and a comment line. Each line must also contain a pseudo random generator seed and an MD5 signature.

\linespread{1.2}
\begin{table}[htb]
\small
\begin{center}
\caption{Instance file.}
\label{instance:table}
\begin{tabular}{lllllllllll}
\hline
ID&Dir&Subdir & n & k & $\rotamax$ & Comment &Seed& MD5\\\hline
0&ex&ex\_0\_0&0&0&2941.15& Max route [...] & 100 & 4d...af\\
1&ex&ex\_10\_5&10&5&2941.15& The size [...] & 101 & e9...10\\
2&ex&ex\_100\_5&100&5&2941.15& Consider [...] & 102 & eb...a2\\
3&ex&ex\_1000\_5&1000&5&2941.15& Instance [...] & 103 & 05...62\\
4&ex&ex\_10000\_5&10000&5&2941.15& Instance [...] & 104 & 93...53\\\hline
\end{tabular}
\end{center}

\end{table}
\linespread{1.5}

\begin{figure}[htb!]
\begin{center}
\includegraphics[width=5cm]{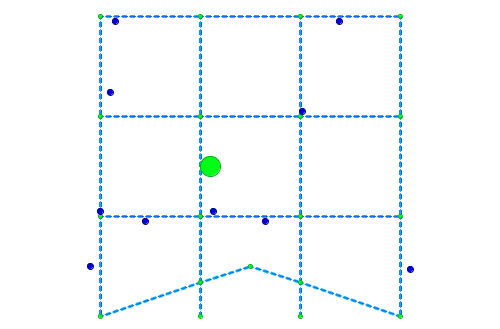}~\hspace{-1cm}
\includegraphics[width=5cm]{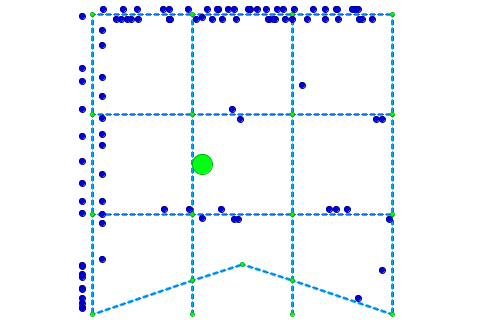}\\
\includegraphics[width=5cm]{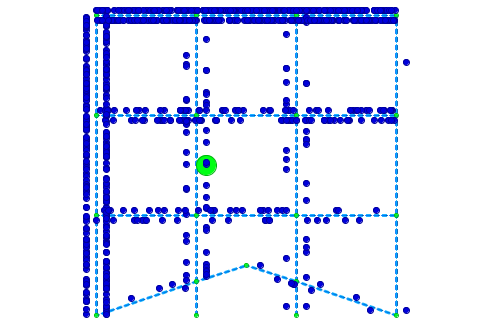}~\hspace{-1cm}
\includegraphics[width=5cm]{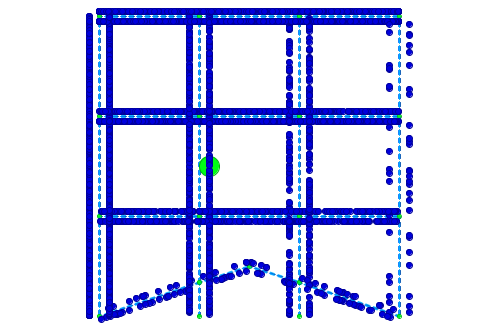}
\end{center}
\caption{Resulting instances with 10, 100, 1,000, and 10,000 deliveries.}
\label{instances}
\end{figure}

The MD5 checksum value is used to ensure the instance identity. The tool will recreate the instances offline and
verify the MD5 signature in the instance file.
The tool will execute the files of Fig.~\ref{table:mod} and Table~\ref{instance:table} and create the instances shown in
Fig.~\ref{instances}.

One can edit the instance file to create new instances for a given model. Once the new instances are created, it is necessary to manually update the MD5 signature. For instance, a new seed will create a new instance with 
another pseudo random sequence.

\begin{figure}[htb!]
\begin{center}
\includegraphics[width=5cm]{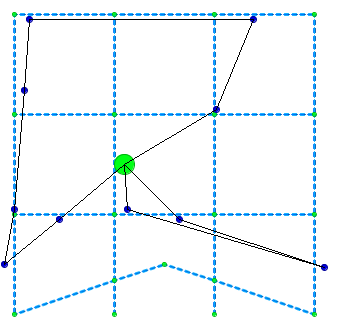}
\caption{Solution to instance with 10 deliveries.}
\label{fig:solution}
\end{center}

\end{figure}

The project site~\citep{vrpbenchsite} provides the source program that parses the instance file.
The program executes a simple swap optimization and saves the route in both a text file and an image file (Fig.~\ref{fig:solution}). 
The purpose is to provide the researcher with a parse to the instance file and a visualization of the solution.

\newcommand{\benchAN}{RWPostVRPB}
\newcommand{\toy}{\textit{Toy}}
\newcommand{\normal}{\textit{Normal}}
\newcommand{\greve}{\textit{OnStrike}}
\newcommand{\natal}{\textit{Christmas}}

\section{Real-world \postVRP benchmark (\benchAN)} \label{sec:bench}
 
In this section, the tool is used to model a real-world 
 mail delivery in the city of \artur, Brazil, namely \benchAN.
%
%
To make the instances as realistic as possible, the authors relied upon
domain expertise from an actual post office in the aforementioned city.


%
%
%

We build the model from a hard copy image instead of using existing street map graphs for the following reasons:
 \textit{(i)} the path on foot may differ from the ones available from graphs which prioritize delivery by vehicles;
 \textit{(ii)} the number of streets in \artur is sufficiently small to allow the manual creation of the graph ($\approx 400$ streets);
and \textit{(iii)} currently, public maps such as OpenStreetMap~\citep{haklay2008openstreetmap} are incomplete, i.e., the city has a large number of streets not covered so far. 

The tool automatically computes corners resulting in a graph with $|V|=2111$ and $|E|=3225$. 
Each vertex is associated with a pixel and
each edge is associated with a straight line between the two edge pixels. The cost of an edge $(u,v)$ is directly proportional to the Euclidean distance between vertices $u$ and $v$ in $\mathbb{R}^2$. 



 Each street is classified using the Region (R), Type (T) and Zone (Z) attributes. 
Each attribute has a corresponding number of levels and each attribute-level 
pair is associated with a multiplicative penalty (\penal) in 
$\mathbb{R}^+$. Table~\ref{table:pen} contains the assignment of attribute, level and penalty based on
expert knowledge.


\linespread{1.2}
\begin{table}[htb!]
\small
\caption{Attribute, level and penalty values.}
\label{table:pen}
\centerline{
\begin{tabular}{llllll}\hline
{Attribute} \!\!&\!\! Level~$1_{(\penal)}$ \!\!&\!\! Level~$2_{(\penal)}$ \!\!&\!\! Level~$3_{(\penal)}$ \!\!&\!\! Level~$4_{(\penal)}$ \\ 
\hline
{Region(\atra)} \!\!&\!\! $\atrana_{(1.0)}$ \!\!&\!\! $\atranb_{(.75)}$ \!\!&\!\!$\atranc_{(.4)}$ \!\!&\!\!$\atrand_{(.2)}$ \\
{Type(\atrb)} \!\!&\!\! $\atrbna_{(1.0)}$ \!\!&\!\! $\atrbnb_{(.75)}$ \!\!&\!\!$\atrbnc_{(.4)}$ \!\!&\!\!$\atrbnd_{(0)}$ \\
{Zone(\atrc)} \!\!&\!\! $\atrcna_{(1.0)}$ \!\!&\!\! $\atrcnb_{(.75)}$ \!\!&\!\!$\atrcnc_{(.4)}$ \!\!&\!\! --- \\\hline
\end{tabular}}
\end{table}
\linespread{1.5}



In the proposed model, streets located in the downtown area have a higher delivery rate per unit length than the ones located in the outskirts.
Such behavior is captured by the Region attribute through four levels: {\it central, peripheral, distant} and {\it isolated.}
The Type attribute has also four levels, namely {\it avenue, street, way} and {\it highway}, while
the Zone attribute may be {\it commercial, mixed}, and {\it residential}.
We used Google Maps as an auxiliary tool to classify streets.
Each one of the \numeroRuas streets received a value in $\atra\times\atrb\times\atrc$, according 
to expert knowledge.

The (non-normalized) probability density $D:\ruas\rightarrow \mathbb{R}^+$ is obtained from the multiplicative penalties.
For example, the $15^{th}$ Avenue is
 $(\atrana,\atrbna,\atrcnb) $, thus
$D(15th)=1\times 1\times 0.7=0.7 $.
On the other hand, \emph{Jasmine} street 
is $(\atrand,\atrbnc,\atrcnc) $, thus
$D(Jasmine)=0.2\times 0.2\times0.4 = 0.16 $.
In this example, a random delivery to the $15^{th}$ avenue is $\frac{0.7}{0.16}$ more probable
than a delivery to $Jasmine$ way by unit length.

The \benchAN~contains 78 instances divided in four groups, \toy, \normal, \greve,~and \natal~(Table~\ref{tab:benchAN}). 
The \toy~set contains 30 instances with a small number of deliveries, and it
may be used to validate algorithms before their use with realistic and larger instances.

\linespread{1.2}
\begin{table}[htb!]
\small
\caption{\benchAN~instances description.}
\vspace{-.5cm}
\label{tab:benchAN}
\begin{center}
\begin{tabular}{lccccccc}\hline
 Set &  \# Instances & \# Deliveries & Length (hrs) & \# Vehicles (max) \\\hline
\toy & 30 & 3 to 5,000 & 6 & 5 to 15\\
\normal & 15 & 10,000 to 14,000 &6 & 30 \\
\greve & 15 & 15,000 to 19,000 & 8 & 30\\
\natal & 18 & 20,000 to 30,000 & 8 & 30\\\hline
\end{tabular}
\end{center}
\end{table}
\linespread{1.5}

The \normal~set contains 15 instances from 10,000 to 14,000 deliveries. \artur city has 
around 50,000 inhabitants and an average of 12,000 daily mail deliveries. 
The normal daily work of a mail carrier has eight hours a day. The mail carrier spends two hours preparing the deliveries inside the post office and six hours 
to complete the deliveries on foot. The post office has around 15 mail carriers to perform the deliveries. 
The number of vehicles is variable in \postVRP, with a maximum value of 30 in \normal~set instances.
 If a feasible solution with 11 mail carriers is found, 
 the post office may assign four mail carriers $(15-11)$ to internal tasks.

\begin{figure}[h!]
	\includegraphics[width=1.0\textwidth]{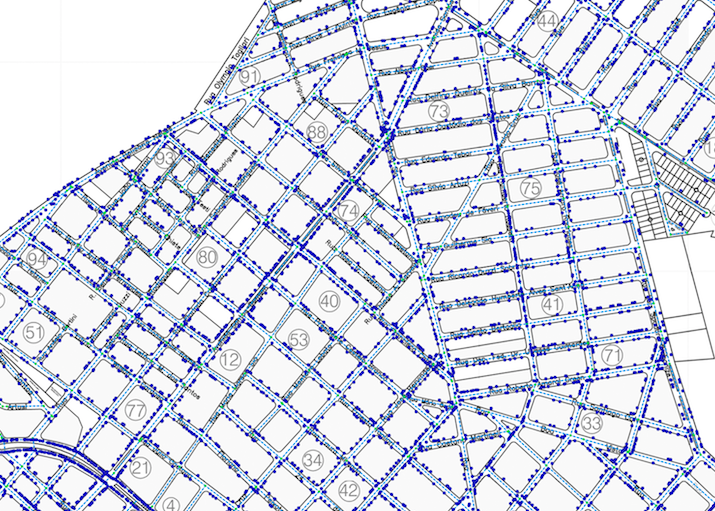}
 	\caption{Part of an instance generated with 10,000 delivery points.}
	\label{part}
\end{figure}

The \textit{OnStrike} and \textit{Christmas} sets may be used to model contingencies.
The \textit{OnStrike} set is similar to \textit{Normal}, but the number of deliveries is larger 
(from 15,000 to 19,000) and the maximum route length is eight hours. The \natal~set models special seasons
 with high delivery rates. The post office often hires extra 
mail carriers for the Christmas season. A feasible solution with 17 mail carriers represents two new hires (15+2).
For all sets, a minimum average route length is desired as well as a minimum variance between the lengths of the routes.
Fig.~\ref{part} shows the distribution of 10,000 delivery points for a chosen area of \artur.

The full set of instances can be downloaded from the project website~\citep{vrpbenchsite}.
The website also includes a pilot sample modeling a section of Manhattan NY. 

\section{Conclusion} \label{sec:conc}


%

In this paper, 
we introduced the \postVRP, a multi-objective VRP variant with a route length constraint. 
The three objectives to be minimized were:
\textit{(i)} the number of vehicles, 
\textit{(ii)} the average length and 
\textit{(iii)} the standard deviation of the length of the routes.

A feasible solution that reduces the number of mail carriers increases the revenues, 
since such mail carriers can be allocated to
other tasks. A solution that reduces the route length also reduces delivery effort, 
which translates to increased profit as well. Reducing the standard 
deviation is also desirable for the sake of fairness of the solution.

By using the tool, we created a benchmark that models a 
problem that comprised the mail delivery on foot in a Brazilian city. The instances were classified into 
four groups:
\textit{(i)} \toy, with up to 5,000 deliveries, 
\textit{(ii)} \normal, with up to 14,000 deliveries, 
\textit{(iii)} \greve, with up to 19,000 deliveries, and 
\textit{(iv)} \natal, with up to 30,000 instances.

The application of the tool allows the generation of arbitrary and large instances in the proposed benchmark by changing the number of deliveries in the instance file. Likewise,
new instances can be created 
by changing the seed of the instance. Additionally, researchers may also model new scenarios. 
To our knowledge, this is the first VRP benchmark with up to 30,000 delivery points that 
models a real-world scenario.

\section{Acknowledgements}

This research was partially supported by FAPESP (Grant Number 2013/23055-5).

\section*{References}

\bibliography{tese}

\begin{thebibliography}{39}
\expandafter\ifx\csname natexlab\endcsname\relax\def\natexlab#1{#1}\fi
\providecommand{\url}[1]{\texttt{#1}}
\providecommand{\href}[2]{#2}
\providecommand{\path}[1]{#1}
\providecommand{\DOIprefix}{doi:}
\providecommand{\ArXivprefix}{arXiv:}
\providecommand{\URLprefix}{URL: }
\providecommand{\Pubmedprefix}{pmid:}
\providecommand{\doi}[1]{\href{http://dx.doi.org/#1}{\path{#1}}}
\providecommand{\Pubmed}[1]{\href{pmid:#1}{\path{#1}}}
\providecommand{\bibinfo}[2]{#2}
\ifx\xfnm\relax \def\xfnm[#1]{\unskip,\space#1}\fi
\bibitem[{Amini et~al.(2010)Amini, Javanshir \&
  Tavakkoli-Moghaddam}]{amini2010pso}
\bibinfo{author}{Amini, S.}, \bibinfo{author}{Javanshir, H.}, \&
  \bibinfo{author}{Tavakkoli-Moghaddam, R.} (\bibinfo{year}{2010}).
\newblock \bibinfo{title}{A pso approach for solving vrptw with real case
  study}.
\newblock {\it \bibinfo{journal}{Int. J. Res. Rev. Appl. Sci}\/},  {\it
  \bibinfo{volume}{4}\/}, \bibinfo{pages}{118--126}.
\bibitem[{Applegate et~al.(2011)Applegate, Bixby, Chvatal \&
  Cook}]{applegate2011travelingP506}
\bibinfo{author}{Applegate, D.~L.}, \bibinfo{author}{Bixby, R.~E.},
  \bibinfo{author}{Chvatal, V.}, \& \bibinfo{author}{Cook, W.~J.}
  (\bibinfo{year}{2011}).
\newblock {\it \bibinfo{title}{The Traveling Salesman Problem: A Computational
  Study}\/}.
\newblock \bibinfo{address}{pp. 506--507}: \bibinfo{publisher}{Princeton
  university press}.
\bibitem[{Augerat et~al.(1995)Augerat, Belenguer, Benavent, Corber{\'a}n,
  Naddef \& Rinaldi}]{augerat1995computational}
\bibinfo{author}{Augerat, P.}, \bibinfo{author}{Belenguer, J.},
  \bibinfo{author}{Benavent, E.}, \bibinfo{author}{Corber{\'a}n, A.},
  \bibinfo{author}{Naddef, D.}, \& \bibinfo{author}{Rinaldi, G.}
  (\bibinfo{year}{1995}).
\newblock {\it \bibinfo{title}{Computational results with a branch and cut code
  for the capacitated vehicle routing problem}\/}.
\newblock \bibinfo{publisher}{IMAG}.
\bibitem[{Bent \& Van~Hentenryck(2004)}]{bent2004two}
\bibinfo{author}{Bent, R.}, \& \bibinfo{author}{Van~Hentenryck, P.}
  (\bibinfo{year}{2004}).
\newblock \bibinfo{title}{A two-stage hybrid local search for the vehicle
  routing problem with time windows}.
\newblock {\it \bibinfo{journal}{Transportation Science}\/},  {\it
  \bibinfo{volume}{38}\/}, \bibinfo{pages}{515--530}.
\bibitem[{du~Buf et~al.(1990)du~Buf, Kardan \& Spann}]{du1990texture}
\bibinfo{author}{du~Buf, J.~H.}, \bibinfo{author}{Kardan, M.}, \&
  \bibinfo{author}{Spann, M.} (\bibinfo{year}{1990}).
\newblock \bibinfo{title}{Texture feature performance for image segmentation}.
\newblock {\it \bibinfo{journal}{Pattern Recognition}\/},  {\it
  \bibinfo{volume}{23}\/}, \bibinfo{pages}{291--309}.
\bibitem[{Burkard et~al.(1997)Burkard, Karisch \& Rendl}]{burkard1997qaplib}
\bibinfo{author}{Burkard, R.~E.}, \bibinfo{author}{Karisch, S.~E.}, \&
  \bibinfo{author}{Rendl, F.} (\bibinfo{year}{1997}).
\newblock \bibinfo{title}{Qaplib--a quadratic assignment problem library}.
\newblock {\it \bibinfo{journal}{Journal of Global optimization}\/},  {\it
  \bibinfo{volume}{10}\/}, \bibinfo{pages}{391--403}.
\bibitem[{Chabrier(2006)}]{chabrier2006vehicle}
\bibinfo{author}{Chabrier, A.} (\bibinfo{year}{2006}).
\newblock \bibinfo{title}{Vehicle routing problem with elementary shortest path
  based column generation}.
\newblock {\it \bibinfo{journal}{Computers \& Operations Research}\/},  {\it
  \bibinfo{volume}{33}\/}, \bibinfo{pages}{2972--2990}.
\bibitem[{Che et~al.(2009)Che, Boyer, Meng, Tarjan, Sheaffer, Lee \&
  Skadron}]{che2009rodinia}
\bibinfo{author}{Che, S.}, \bibinfo{author}{Boyer, M.}, \bibinfo{author}{Meng,
  J.}, \bibinfo{author}{Tarjan, D.}, \bibinfo{author}{Sheaffer, J.~W.},
  \bibinfo{author}{Lee, S.-H.}, \& \bibinfo{author}{Skadron, K.}
  (\bibinfo{year}{2009}).
\newblock \bibinfo{title}{Rodinia: A benchmark suite for heterogeneous
  computing}.
\newblock In {\it \bibinfo{booktitle}{Workload Characterization, 2009. IISWC
  2009. IEEE International Symposium on}\/} (pp. \bibinfo{pages}{44--54}).
\newblock \bibinfo{organization}{IEEE}.
\bibitem[{Christofides(1976)}]{christofides1976vehicle}
\bibinfo{author}{Christofides, N.} (\bibinfo{year}{1976}).
\newblock \bibinfo{title}{The vehicle routing problem}.
\newblock {\it \bibinfo{journal}{Revue fran{\c{c}}aise d'automatique,
  d'informatique et de recherche op{\'e}rationnelle. Recherche
  Op{\'e}rationnelle}\/},  {\it \bibinfo{volume}{10}\/},
  \bibinfo{pages}{55--70}.
\bibitem[{Christofides(1979)}]{christofides1979vehicle}
\bibinfo{author}{Christofides, N.} (\bibinfo{year}{1979}).
\newblock \bibinfo{title}{The vehicle routing problem. combinatorial
  optimization. christofides n., mingozzi a., toth p., sandi c.(eds) j}.
\bibitem[{Christofides \& Eilon(1969)}]{christofides1969algorithm}
\bibinfo{author}{Christofides, N.}, \& \bibinfo{author}{Eilon, S.}
  (\bibinfo{year}{1969}).
\newblock \bibinfo{title}{An algorithm for the vehicle-dispatching problem}.
\newblock {\it \bibinfo{journal}{Or}\/},  (pp. \bibinfo{pages}{309--318}).
\bibitem[{Contardo \& Martinelli(2014)}]{contardo2014new}
\bibinfo{author}{Contardo, C.}, \& \bibinfo{author}{Martinelli, R.}
  (\bibinfo{year}{2014}).
\newblock \bibinfo{title}{A new exact algorithm for the multi-depot vehicle
  routing problem under capacity and route length constraints}.
\newblock {\it \bibinfo{journal}{Discrete Optimization}\/},  {\it
  \bibinfo{volume}{12}\/}, \bibinfo{pages}{129--146}.
\bibitem[{Cook(2016)}]{waterloosite}
\bibinfo{author}{Cook, W.} (\bibinfo{year}{2016}).
\newblock \bibinfo{title}{Traveling salesman problem site}.
\newblock \bibinfo{howpublished}{(last check 01/30/2016)
  \url{http://www.math.uwaterloo.ca/tsp/d15sol/}}.
\bibitem[{Correia et~al.(2015)Correia, Hohendorff, Gaspar \&
  Schiozer}]{correia2015unisim}
\bibinfo{author}{Correia, M.}, \bibinfo{author}{Hohendorff, J.},
  \bibinfo{author}{Gaspar, A. T. F.~S.}, \& \bibinfo{author}{Schiozer, D.}
  (\bibinfo{year}{2015}).
\newblock \bibinfo{title}{Unisim-ii-d: Benchmark case proposal based on a
  carbonate reservoir}.
\newblock In {\it \bibinfo{booktitle}{SPE Latin American and Caribbean
  Petroleum Engineering Conference}\/}.
\newblock \bibinfo{organization}{Society of Petroleum Engineers}.
\bibitem[{Dantzig \& Ramser(1959)}]{dantzig1959truck}
\bibinfo{author}{Dantzig, G.~B.}, \& \bibinfo{author}{Ramser, J.~H.}
  (\bibinfo{year}{1959}).
\newblock \bibinfo{title}{The truck dispatching problem}.
\newblock {\it \bibinfo{journal}{Management science}\/},  {\it
  \bibinfo{volume}{6}\/}, \bibinfo{pages}{80--91}.
\bibitem[{Espinoza(2006)}]{espinoza2006linear}
\bibinfo{author}{Espinoza, D.~G.} (\bibinfo{year}{2006}).
\newblock {\it \bibinfo{title}{On linear programming, integer programming and
  cutting planes}\/}.
\newblock Ph.D. thesis Georgia Institute of Technology.
\bibitem[{Fisher(1994)}]{fisher1994optimal}
\bibinfo{author}{Fisher, M.~L.} (\bibinfo{year}{1994}).
\newblock \bibinfo{title}{Optimal solution of vehicle routing problems using
  minimum k-trees}.
\newblock {\it \bibinfo{journal}{Operations research}\/},  {\it
  \bibinfo{volume}{42}\/}, \bibinfo{pages}{626--642}.
\bibitem[{Fukasawa et~al.(2006)Fukasawa, Longo, Lysgaard, de~Arag{\~a}o, Reis,
  Uchoa \& Werneck}]{fukasawa2006robust}
\bibinfo{author}{Fukasawa, R.}, \bibinfo{author}{Longo, H.},
  \bibinfo{author}{Lysgaard, J.}, \bibinfo{author}{de~Arag{\~a}o, M.~P.},
  \bibinfo{author}{Reis, M.}, \bibinfo{author}{Uchoa, E.}, \&
  \bibinfo{author}{Werneck, R.~F.} (\bibinfo{year}{2006}).
\newblock \bibinfo{title}{Robust branch-and-cut-and-price for the capacitated
  vehicle routing problem}.
\newblock {\it \bibinfo{journal}{Mathematical programming}\/},  {\it
  \bibinfo{volume}{106}\/}, \bibinfo{pages}{491--511}.
\bibitem[{Gehring \& Homberger(1999)}]{gehring1999parallel}
\bibinfo{author}{Gehring, H.}, \& \bibinfo{author}{Homberger, J.}
  (\bibinfo{year}{1999}).
\newblock \bibinfo{title}{A parallel hybrid evolutionary metaheuristic for the
  vehicle routing problem with time windows}.
\newblock In {\it \bibinfo{booktitle}{Proceedings of EUROGEN99}\/} (pp.
  \bibinfo{pages}{57--64}).
\newblock \bibinfo{organization}{Springer Berlin} volume~\bibinfo{volume}{2}.
\bibitem[{Golden et~al.(1998)Golden, Wasil, Kelly \& Chao}]{golden1998impact}
\bibinfo{author}{Golden, B.~L.}, \bibinfo{author}{Wasil, E.~A.},
  \bibinfo{author}{Kelly, J.~P.}, \& \bibinfo{author}{Chao, I.-M.}
  (\bibinfo{year}{1998}).
\newblock \bibinfo{title}{The impact of metaheuristics on solving the vehicle
  routing problem: algorithms, problem sets, and computational results}.
\newblock In {\it \bibinfo{booktitle}{Fleet management and logistics}\/} (pp.
  \bibinfo{pages}{33--56}).
\newblock \bibinfo{publisher}{Springer}.
\bibitem[{Haklay \& Weber(2008)}]{haklay2008openstreetmap}
\bibinfo{author}{Haklay, M.}, \& \bibinfo{author}{Weber, P.}
  (\bibinfo{year}{2008}).
\newblock \bibinfo{title}{Openstreetmap: User-generated street maps}.
\newblock {\it \bibinfo{journal}{IEEE Pervasive Computing}\/},  {\it
  \bibinfo{volume}{7}\/}, \bibinfo{pages}{12--18}.
\bibitem[{Huang et~al.(2007)Huang, Ramesh, Berg \&
  Learned-Miller}]{huang2007labeled}
\bibinfo{author}{Huang, G.~B.}, \bibinfo{author}{Ramesh, M.},
  \bibinfo{author}{Berg, T.}, \& \bibinfo{author}{Learned-Miller, E.}
  (\bibinfo{year}{2007}).
\newblock {\it \bibinfo{title}{Labeled faces in the wild: A database for
  studying face recognition in unconstrained environments}\/}.
\newblock \bibinfo{type}{Technical Report} Technical Report 07-49, University
  of Massachusetts, Amherst.
\bibitem[{Jawarneh \& Abdullah(2015)}]{jawarneh2015sequential}
\bibinfo{author}{Jawarneh, S.}, \& \bibinfo{author}{Abdullah, S.}
  (\bibinfo{year}{2015}).
\newblock \bibinfo{title}{Sequential insertion heuristic with adaptive bee
  colony optimisation algorithm for vehicle routing problem with time windows}.
\newblock {\it \bibinfo{journal}{PLOS ONE}\/},  {\it \bibinfo{volume}{10}\/},
  \bibinfo{pages}{1--23}. \URLprefix
  \url{https://doi.org/10.1371/journal.pone.0130224}.
  \DOIprefix\doi{10.1371/journal.pone.0130224}.
\bibitem[{Johnson(2002)}]{johnson2002theoretician}
\bibinfo{author}{Johnson, D.~S.} (\bibinfo{year}{2002}).
\newblock \bibinfo{title}{A theoretician's guide to the experimental analysis
  of algorithms}.
\newblock {\it \bibinfo{journal}{Data structures, near neighbor searches, and
  methodology: fifth and sixth DIMACS implementation challenges}\/},  {\it
  \bibinfo{volume}{59}\/}, \bibinfo{pages}{215--250}.
\bibitem[{Jorion(1997)}]{jorion1997value}
\bibinfo{author}{Jorion, P.} (\bibinfo{year}{1997}).
\newblock {\it \bibinfo{title}{Value at risk: the new benchmark for controlling
  market risk}\/}.
\newblock \bibinfo{publisher}{Irwin Professional Pub.}
\bibitem[{Kallehauge et~al.(2005)Kallehauge, Larsen, Madsen \&
  Solomon}]{kallehauge2005vehicle}
\bibinfo{author}{Kallehauge, B.}, \bibinfo{author}{Larsen, J.},
  \bibinfo{author}{Madsen, O.~B.}, \& \bibinfo{author}{Solomon, M.~M.}
  (\bibinfo{year}{2005}).
\newblock {\it \bibinfo{title}{Vehicle routing problem with time windows}\/}.
\newblock \bibinfo{publisher}{Springer}.
\bibitem[{Kohl et~al.(1999)Kohl, Desrosiers, Madsen, Solomon \&
  Soumis}]{kohl19992}
\bibinfo{author}{Kohl, N.}, \bibinfo{author}{Desrosiers, J.},
  \bibinfo{author}{Madsen, O.~B.}, \bibinfo{author}{Solomon, M.~M.}, \&
  \bibinfo{author}{Soumis, F.} (\bibinfo{year}{1999}).
\newblock \bibinfo{title}{2-path cuts for the vehicle routing problem with time
  windows}.
\newblock {\it \bibinfo{journal}{Transportation Science}\/},  {\it
  \bibinfo{volume}{33}\/}, \bibinfo{pages}{101--116}.
\bibitem[{Kolisch \& Sprecher(1997)}]{kolisch1997psplib}
\bibinfo{author}{Kolisch, R.}, \& \bibinfo{author}{Sprecher, A.}
  (\bibinfo{year}{1997}).
\newblock \bibinfo{title}{Psplib-a project scheduling problem library: Or
  software-orsep operations research software exchange program}.
\newblock {\it \bibinfo{journal}{European journal of operational research}\/},
  {\it \bibinfo{volume}{96}\/}, \bibinfo{pages}{205--216}.
\bibitem[{Krizhevsky et~al.(2012)Krizhevsky, Sutskever \&
  Hinton}]{krizhevsky2012imagenet}
\bibinfo{author}{Krizhevsky, A.}, \bibinfo{author}{Sutskever, I.}, \&
  \bibinfo{author}{Hinton, G.~E.} (\bibinfo{year}{2012}).
\newblock \bibinfo{title}{Imagenet classification with deep convolutional
  neural networks}.
\newblock In {\it \bibinfo{booktitle}{Advances in neural information processing
  systems}\/} (pp. \bibinfo{pages}{1097--1105}).
\bibitem[{Li et~al.(2005)Li, Golden \& Wasil}]{li2005very}
\bibinfo{author}{Li, F.}, \bibinfo{author}{Golden, B.}, \&
  \bibinfo{author}{Wasil, E.} (\bibinfo{year}{2005}).
\newblock \bibinfo{title}{Very large-scale vehicle routing: new test problems,
  algorithms, and results}.
\newblock {\it \bibinfo{journal}{Computers \& Operations Research}\/},  {\it
  \bibinfo{volume}{32}\/}, \bibinfo{pages}{1165--1179}.
\bibitem[{Meira et~al.(2017)Meira, Zeni, Menzori \& Martins}]{vrpbenchsite}
\bibinfo{author}{Meira, L. A.~A.}, \bibinfo{author}{Zeni, G.~A.},
  \bibinfo{author}{Menzori, M.}, \& \bibinfo{author}{Martins, P.~S.}
  (\bibinfo{year}{2017}).
\newblock \bibinfo{title}{Postvrp project site (last checked sep--2017)}.
\newblock \URLprefix \url{http://www.ft.unicamp.br/~meira/postvrp}.
\bibitem[{Pecin et~al.(2014)Pecin, Pessoa, Poggi \& Uchoa}]{pecin2014improved}
\bibinfo{author}{Pecin, D.}, \bibinfo{author}{Pessoa, A.},
  \bibinfo{author}{Poggi, M.}, \& \bibinfo{author}{Uchoa, E.}
  (\bibinfo{year}{2014}).
\newblock \bibinfo{title}{Improved branch-cut-and-price for capacitated vehicle
  routing}.
\newblock In {\it \bibinfo{booktitle}{Integer programming and combinatorial
  optimization}\/} (pp. \bibinfo{pages}{393--403}).
\newblock \bibinfo{publisher}{Springer}.
\bibitem[{Reinelt(1991)}]{reinelt1991tsplib}
\bibinfo{author}{Reinelt, G.} (\bibinfo{year}{1991}).
\newblock \bibinfo{title}{Tsplib---a traveling salesman problem library}.
\newblock {\it \bibinfo{journal}{ORSA journal on computing}\/},  {\it
  \bibinfo{volume}{3}\/}, \bibinfo{pages}{376--384}.
\bibitem[{Reinelt(1995)}]{reinelt1995tsplib95}
\bibinfo{author}{Reinelt, G.} (\bibinfo{year}{1995}).
\newblock {\it \bibinfo{title}{Tsplib95}\/}.
\newblock \bibinfo{type}{Technical Report} Interdisziplin{\"a}res Zentrum
  f{\"u}r Wissenschaftliches Rechnen (IWR), Heidelberg.
\bibitem[{Renaud et~al.(1996)Renaud, Laporte \& Boctor}]{renaud1996tabu}
\bibinfo{author}{Renaud, J.}, \bibinfo{author}{Laporte, G.}, \&
  \bibinfo{author}{Boctor, F.~F.} (\bibinfo{year}{1996}).
\newblock \bibinfo{title}{A tabu search heuristic for the multi-depot vehicle
  routing problem}.
\newblock {\it \bibinfo{journal}{Computers \& Operations Research}\/},  {\it
  \bibinfo{volume}{23}\/}, \bibinfo{pages}{229--235}.
\bibitem[{Solomon(1987)}]{solomon1987algorithms}
\bibinfo{author}{Solomon, M.~M.} (\bibinfo{year}{1987}).
\newblock \bibinfo{title}{Algorithms for the vehicle routing and scheduling
  problems with time window constraints}.
\newblock {\it \bibinfo{journal}{Operations research}\/},  {\it
  \bibinfo{volume}{35}\/}, \bibinfo{pages}{254--265}.
\bibitem[{Tol(2002)}]{tol2002estimates}
\bibinfo{author}{Tol, R.~S.} (\bibinfo{year}{2002}).
\newblock \bibinfo{title}{Estimates of the damage costs of climate change. part
  1: Benchmark estimates}.
\newblock {\it \bibinfo{journal}{Environmental and Resource Economics}\/},
  {\it \bibinfo{volume}{21}\/}, \bibinfo{pages}{47--73}.
\bibitem[{Uchoa et~al.(2017{\natexlab{a}})Uchoa, Pecin, Pessoa, Poggi,
  Subramanian \& Vidal}]{CVRPLIB}
\bibinfo{author}{Uchoa, E.}, \bibinfo{author}{Pecin, D.},
  \bibinfo{author}{Pessoa, A.}, \bibinfo{author}{Poggi, M.},
  \bibinfo{author}{Subramanian, A.}, \& \bibinfo{author}{Vidal, T.}
  (\bibinfo{year}{2017}{\natexlab{a}}).
\newblock \bibinfo{title}{Cvrp library site (last checked sep--2017)}.
\newblock \URLprefix \url{http://www.galgos.inf.puc-rio.br/vrp/}.
\bibitem[{Uchoa et~al.(2017{\natexlab{b}})Uchoa, Pecin, Pessoa, Poggi, Vidal \&
  Subramanian}]{uchoa2017}
\bibinfo{author}{Uchoa, E.}, \bibinfo{author}{Pecin, D.},
  \bibinfo{author}{Pessoa, A.}, \bibinfo{author}{Poggi, M.},
  \bibinfo{author}{Vidal, T.}, \& \bibinfo{author}{Subramanian, A.}
  (\bibinfo{year}{2017}{\natexlab{b}}).
\newblock \bibinfo{title}{New benchmark instances for the capacitated vehicle
  routing problem}.
\newblock {\it \bibinfo{journal}{European Journal of Operational Research}\/},
  {\it \bibinfo{volume}{257}\/}, \bibinfo{pages}{845 -- 858}. \URLprefix
  \url{http://www.sciencedirect.com/science/article/pii/S0377221716306270}.
  \DOIprefix\doi{https://doi.org/10.1016/j.ejor.2016.08.012}.

\end{thebibliography}

\end{document}